\relax
\documentclass[preprint,3p]{elsarticle}
\usepackage{times}  
\usepackage{helvet} 
\usepackage{courier}  
\usepackage[hyphens]{url}  
\usepackage{graphicx} 
\usepackage{graphicx}  
\usepackage{natbib}  
\usepackage{caption} 
\usepackage{algorithm}
\usepackage{algorithmic}

\usepackage{graphicx}  
\usepackage{float}  
\usepackage{subfigure}  
\usepackage{amsmath}
\usepackage{amsfonts}
\usepackage[table,xcdraw]{xcolor}
\usepackage{booktabs}
\usepackage{tabularx}
\usepackage{multirow}
\usepackage{array}
\usepackage{geometry}
\usepackage[utf8]{inputenc}
\usepackage{indentfirst}
\usepackage{wrapfig}
\usepackage{graphicx}
\usepackage{multicol}
\usepackage{balance}

\begin{document}
\captionsetup[figure]{labelfont={bf},name={Fig.},labelsep=period}

\title{Channel Pruning Guided by Spatial and Channel Attention for DNNs in Intelligent Edge Computing}

\begin{abstract}
\small{Deep Neural Networks (DNNs) have achieved remarkable success in many computer vision tasks recently, but the huge number of parameters and the high computation overhead hinder their deployments on resource-constrained edge devices. It is worth noting that channel pruning is an effective approach for compressing DNN models. A critical challenge is to determine which channels are to be removed, so that the model accuracy will not be negatively affected. In this paper, we first propose Spatial and Channel Attention (SCA), a new attention module combining both spatial and channel attention that respectively focuses on ``where'' and ``what'' are the most informative parts. Guided by the scale values generated by SCA for measuring channel importance, we further propose a new channel pruning approach called Channel Pruning guided by Spatial and Channel Attention (CPSCA). Experimental results indicate that SCA achieves the best inference accuracy, while incurring negligibly extra resource consumption, compared to other state-of-the-art attention modules. Our evaluation on two benchmark datasets shows that, with the guidance of SCA, our CPSCA approach achieves higher inference accuracy than other state-of-the-art pruning methods under the same pruning ratios.}
\end{abstract}


\author[a]{Mengran Liu}
\author[a,b]{Weiwei Fang\corref{cor1}}
\ead{fangww@bjtu.edu.cn}
\author[a]{Xiaodong Ma}
\author[a]{Wenyuan Xu}
\author[c]{Naixue Xiong}
\author[d]{Yi Ding}

\cortext[cor1]{Corresponding author}
\address[a]{School of Computer and Information Technology, Beijing Jiaotong University, Beijing 100044, China}
\address[b]{Key Laboratory of Industrial Internet of Things \& Networked Control, Ministry of Education, Chongqing 400065, China}
\address[c]{College of Intelligence and Computing, Tianjin University, Tianjin 300350, China}
\address[d]{School of Information, Beijing Wuzi University, Beijing 101149, China}

\captionsetup{belowskip=0pt} 


\maketitle

\section{Introduction}
Intelligent edge, which refers to the combination of edge computing with artificial intelligence (AI), machine learning, big data and other technologies, is regarded as a crucial step in the new revolution of the Internet of things (IoT) domain. Intelligent edge collects, stores, and processes data in real time by edge devices. It can not only make full use of the computing power of edge devices, but also reduce the bandwidth consumption, improve the response speed, and minimize the risk of data privacy leakage, as compared with traditional cloud computing. However, there exist many challenges for deploying AI techniques on edge devices. As the most popular AI technique, DNNs (Deep Neural Networks) have achieved remarkable success in many application scenarios, ranging from the initial handwriting recognition \cite{plamondon2000online} to image classification \cite{wan2014deep}, object detection \cite{girshick2015fast}, and visual tracking \cite{wang2013learning}. DNNs have displaced conventional computer vision approaches to a great extent, since they can provide near-human or even better-than-human accuracy in practice \cite{He2016}. However, DNNs are known to be both compute-intensive and memory-intensive. For example, VGG16 has 138M weights and requires 15.5G multiply-and-accumulates for one input image, and ResNet50 has 25.5M weights and requires 3.9G multiply-and-accumulates per image \cite{He2016}. Thus, most DNN models currently are difficult to be deployed on resource-constrained Internet-of-Things (IoT) devices and in performance-demanding edge-computing applications. A critical challenge is how to compress the DNN models to reduce computational requirements and resource consumption without negatively affecting their accuracy \cite{Zhang2015,Huang2017}.

In this context, DNN compression techniques have been intensively studied, e.g., parameter pruning \cite{li2016pruning}, low-rank factorization \cite{Grachev2019}, weight quantization \cite{han2015deep}, knowledge distillation \cite{Zhang2020}, etc. Among them, the pruning-based methods aim to compress DNN models and accelerate DNN inference by removing redundancy in structures and parameters. Except for early works on non-structured weight pruning \cite{han2015deep}, the structured pruning approach represented by channel pruning becomes more popular in recent years, since it does not require using specialized software and hardware. The basic idea of channel pruning is to reduce the number of input/output channels in the convolutional layer with negligible performance degradation \cite{he2017channel,luo2017thinet}. A key step in channel pruning is to measure the importance of each channel, which determines if the channel could be removed or not. Early studies only take the independent contribution of each channel to reconstruction loss into consideration \cite{luo2017thinet,yu2018nisp,he2017channel}, and neglect the impact of correlations between neighboring channels to inference performance.Besides, the tradeoff between accuracy and pruned ratio is a noteworthy problem. In order to achieve a better balance between the pruned ratio and accuracy, the work in \cite{2020Deep} proposed a efficient approach to channel pruning, based on the genetic algorithm and sparse learning, and another work \cite{wang2020network} proposed a scheme of network compression from the perspective of multi-objective evolution. However, the accuracy and pruned ratio of the existing methods need to be improved.

Attention mechanism is a good alternative for measuring the important level of channels and the inter-channel relationship of features \cite{woo2018cbam}. It is inspired by human perception, in which our central nervous system tends to focus on salient parts from a series of partial sense-organ input so as to capture perceptual information better \cite{xu2015show}. Attention not only tells us where to focus, but also helps us to improve the representation of subject of interest. This mechanism initially attracts widespread interest in the field of natural language processing \cite{bahdanau2014neural}, and then achieves a lot of promising results in emerging computer vision applications. For image classification tasks, there have been several attempts to incorporate attention processing to improve the inference performance of DNN models. Typical attention modules for image classification include Residual Attention Network \cite{wang2017residual}, SENet \cite{hu2018squeeze}, BAM \cite{park2018bam}, CBAM \cite{woo2018cbam} and SGE \cite{li2019spatial}.

In this paper, we propose a new channel pruning approach called Channel Pruning guided by Spatial and Channel Attention (CPSCA), in which we combine both spatial attention and channel attention together to determine the channel importance and guide the pruning operation. At first, we design a simple yet effective Spatial and Channel Attention module (SCA). This SCA module can emphasize useful features and suppress useless features along both the spatial and channel dimensions, so as to boost representation power of current DNN model. Meanwhile, it can generate a scale value for each individual channel to measure its important level to the classification task. Using this scale value, we develop a CPSCA pruning approach to prune redundant and unimportant channels from the original DNN model, so as to well reduce resource consumption in terms of computation and storage while incurring negligible accuracy loss. On the one hand, compared with the existing attention designs (e.g., \cite{hu2018squeeze,woo2018cbam,li2019spatial}), SCA adopts pooling and group normalization to improve the performance of spatial and channel attention submodules, respectively. On the other hand, compared with the exiting pruning techniques (e.g., \cite{li2016pruning,liu2017learning,song2018channel}), CPSCA is more capable to classify and prune unimportant channels so as to well preserve the inference accuracy.

The main contributions of this work are summarized as follows:
\begin{itemize}
\item We design a new light-weight attention module SCA, which combines both spatial and channel attention as a whole. It can not only be inserted into the DNN model in a plug-and-play fashion to enhance representation power, but also output the scale values as a more reliable measure of importance for the channels of the DNN model.

\item Guided by SCA, we further propose a new attention-based channel pruning approach CPSCA. This approach measures the important level of channels based on the attention statistics from the SCA module. It then removes the unimportant channels from the DNN model, and retrains (fine-tunes) the model for loss recovery.

\item We verify the effectiveness of the optimal structure design of SCA through ablation studies, and demonstrate the superiority of this module by comparing it with the state-of-the-art attention modules using CIFAR datasets and VGG/ResNet models.

\item We conduct extensive experiments on CIFAR-10 and CIFAR-100 datasets, and the results show that our CPSCA pruning approach achieves higher inference accuracy than other state-of-the-art pruning methods under the same pruning ratios.
\end{itemize}

The remainder of this paper is organized as follows. In Section 2, we briefly review related works. Section 3 introduces the proposed SCA attention module and the CPSCA approach. In Sections 4, we evaluate the performance of SCA and CPSCA using the standard datasets CIFAR-10 and CIFAR-100 on image classification. Section 5 finally concludes the paper.
\section{Related Work}

\textbf{Pruning approaches.} The pruning approaches for DNNs can be generally classified as two categories, i.e., non-structured pruning and structured pruning. Early studies \cite{han2015deep,han2015learning,luo2017entropy,zhang2018systematic} are mainly based on weight pruning, resulting in non-structured sparsity in the pruned model. The runtime acceleration is difficult to be achieved because of irregular memory access \cite{wen2016learning}, unless using specialized hardware and libraries. Pioneered by \cite{wen2016learning,he2017channel}, structured pruning overcomes this problem by removing whole filters or channels and producing a non-sparse compressed model \cite{wen2016learning,luo2017entropy,min20182pfpce}. Structured pruning can be classified into four types according to pruning granularities, including layer pruning, filter pruning, channel pruning and kernel pruning. It is noteworthy that channel pruning and filter pruning are correlated, because pruning the channel of current layer will cause the corresponding filter of the upper layer to be removed \cite{he2017channel}. In this paper, we focus on channel pruning, which targets at removing a certain number of channels and the relevant filters to compress DNN models. $\ell_{1}$-norm based work \cite{li2016pruning} used the $\ell_{1}$-norm of filters as the pruning criterion. Network Slimming \cite{liu2017learning} adopted the Batch Normalization (BN) layers to scale different channels, and identified the channels with relatively smaller scale factors as the unimportant ones to be pruned. Some studies proposed to prune the channels that have the smallest impact on the feature reconstruction error between the original model and the pruned model. The work in \cite{luo2017thinet} proposed a greedy search based method to minimize the reconstruction error, and another work \cite{he2017channel} retained the representative channels by solving a lasso regression problem about the reconstruction error. Both of them only considered local statistics of two consecutive layers, i.e., prune one layer to minimize the next layer's reconstruction error. Considering the effect of potential error propagation in the entire network, NISP \cite{yu2018nisp} proposed to minimize the reconstruction errors using the global importance scores propagated from the second-to-last layer before classification. It must be noted that a good metric for channel pruning should take not only the channel importance from a global view but also the correlation between different channels \cite{yu2018nisp,hu2018squeeze}.

\textbf{Attention mechanisms.} The attention mechanism can effectively help to improve the classification performance of DNNs \cite{li2019selective}, by enhancing the representation of feature map with more important information and suppressing the interference of unnecessary information \cite{larochelle2010learning,mnih2014recurrent}. Attention has been widely used in recent applications, e.g., neural machine translation \cite{bahdanau2014neural}, image captioning \cite{you2016image}, object detection \cite{hu2018relation}, and generative modeling \cite{zhang2019self}. To improve the performance of image classification, SENet (Squeeze-and-Excitation Networks) \cite{hu2018squeeze} proposed a light-weight gating mechanism to exploit the channe{}l-wise relationships. It is actually an attention mechanism applied along the channel dimension, but neglects the importance of the spatial dimension. The SGE (Spatial Group-wise Enhance) module \cite{li2019spatial} is designed to enhance the ability of each of its feature groups to express different semantics while suppressing possible noise and interference. It is essentially a spatial attention module that misses the spatial attention. Both CBAM (Convolutional Block Attention Module) \cite{woo2018cbam} and BAM (Bottleneck Attention Module) \cite{park2018bam} exploit both spatial-wise and channel-wise attention, and verify that combining both is superior to using either of them. The structure design of these two modules are different. CBAM sequentially apply channel and spatial attention, while BAM computes the two attentions in a simultaneous way.

\textbf{Attention-based pruning methods.} In recent years, attention mechanisms have also been introduced for improving the performance of model pruning. For example, the work in \cite{song2018channel} proposed to apply the SENet model to evaluate channel importance, so that the redundant channels with least importance can be identified and removed. However, the limitation of SENet itself makes the scale value generated by this model can not fully reflect the channel importance and improve the pruning performance. PCAS \cite{yamamoto2018pcas} designed a new attention model and evaluated the importance of channels based on attention statistics. Actually, the module in PCAS is only a channel attention module, and has two fully connected layers that incur additional overhead and complexity. Moreover, the operations of dimensionality reduction bring side effect \cite{szegedy2016rethinking,gao2019global} on channel attention prediction. To address these problems, we propose a new attention module that exploits both spatial and channel-wise attention based on an efficient structure design, and verify its superior performance over state-of-the-art solutions in the process of channel pruning.

\section{Our Proposed CPSCA Methodology}

In this section, we first present an overview of our CPSCA approach. Next, we introduce the structural composition of the SCA attention module. Finally, we propose the CPSCA algorithm that prunes DNN models with the guidance of SCA.

\subsection{Overview of Our CPSCA Approach}

\begin{figure}[htb]
\centering
\includegraphics[scale=1.2]{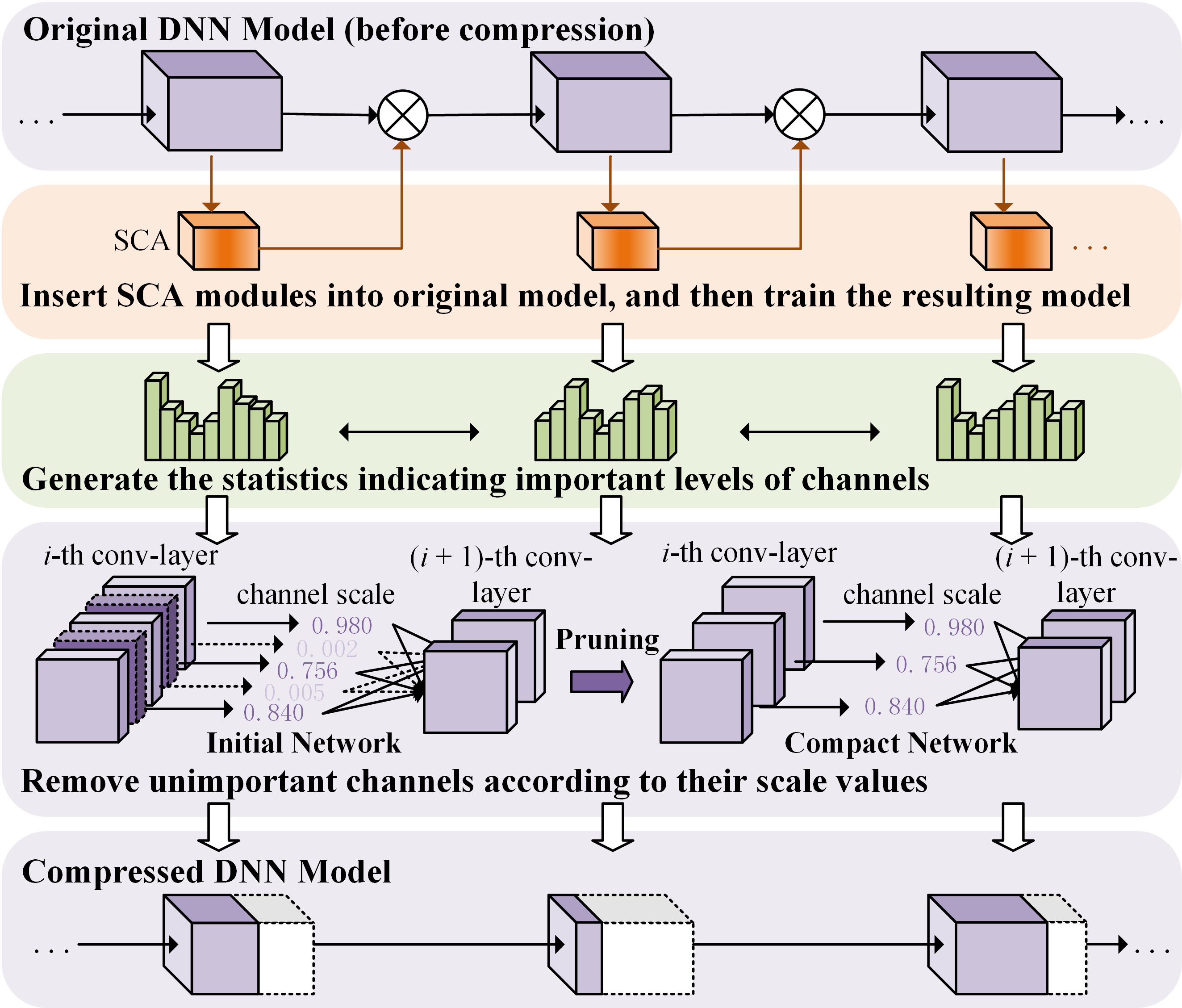}
\caption{Overview of the CPSCA approach.}
\label{fig1}
\end{figure}

Fig. \ref{fig1} depicts the overview of our CPSCA approach. Firstly, we insert our SCA modules which can reflect the importance of channels into the original network, and then train the resulting network. According to the statistical channel scales generated by SCA and the pre-defined channel pruning ratios, a certain number of channels with least importance are identified. After removing all inserted SCA modules, we then prune the network by removing the identified channels as well as the filters corresponding to these channels. At last, we finetune the pruned network to recover the accuracy loss caused by pruning.

\begin{figure}[htb]
\centering
\includegraphics[scale=1]{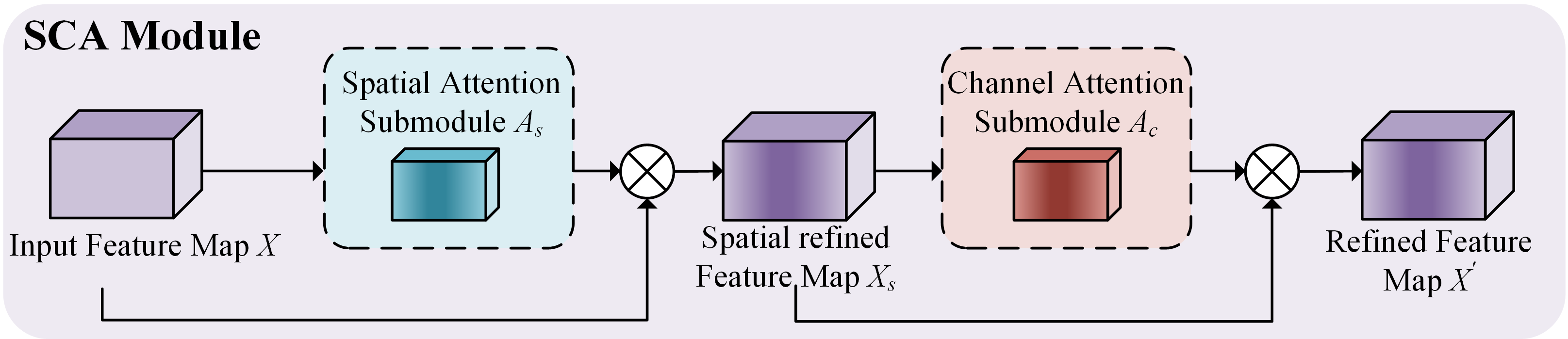}
 \caption{Overall structure of the SCA module.}
\label{fig2}
\end{figure}

Fig. \ref{fig2} shows the overall structure design of our SCA (Spatial and Channel Attention) module. In fact, if we only employ spatial attention, the information in the channel dimension will be ignored, as it treats the features in different channels equally. Similarly, if we only employ channel attention, the information inside of channels will also be ignored. Thus, we believe the combination of spatial and channel attention modules as a whole module will achieve higher performance.

\subsection{SCA Module}

Given the input feature map $X \in \mathbb{R}^{C \times H \times W}$, the spatial attention submodule firstly infers the spatial attention map $A_s \in \mathbb{R}^{1 \times H \times W}$, then we can obtain the spatial refined feature map $X_s \in \mathbb{R}^{C \times H \times W}$ by:

\begin{equation}
\label{eq1}
    X_s = X \otimes A_s(X),
\end{equation}

\noindent where $\otimes$ denotes element-wise multiplication. Based on $X_s$, the channel attention submodule further deduces the channel attention map $A_c \in \mathbb{R}^{C \times 1 \times 1}$, and then generates the final refined feature map $X^{'} \in \mathbb{R}^{C \times H \times W}$ by:

\begin{equation}
\label{eq2}
    X^{'} = X_s \otimes A_c(X_s).
\end{equation}

The computation process of spatial and channel attention submodules are introduced in the following subsections.

\subsubsection{Spatial Attention Submodule}

The spatial attention submodule focuses on ‘where’ are the informative parts, and pays differentiated attention to different positions on the feature map. For a feature map $X \in \mathbb{R}^{C \times H \times W}$, the spatial attention map $A_s \in \mathbb{R}^{1 \times H \times W}$ is corresponding to a $H \times W$ matrix, in which each position denotes a weight corresponding to a pixel of original feature map.

\begin{figure*}[htb]
\centering
\includegraphics[scale=1.35]{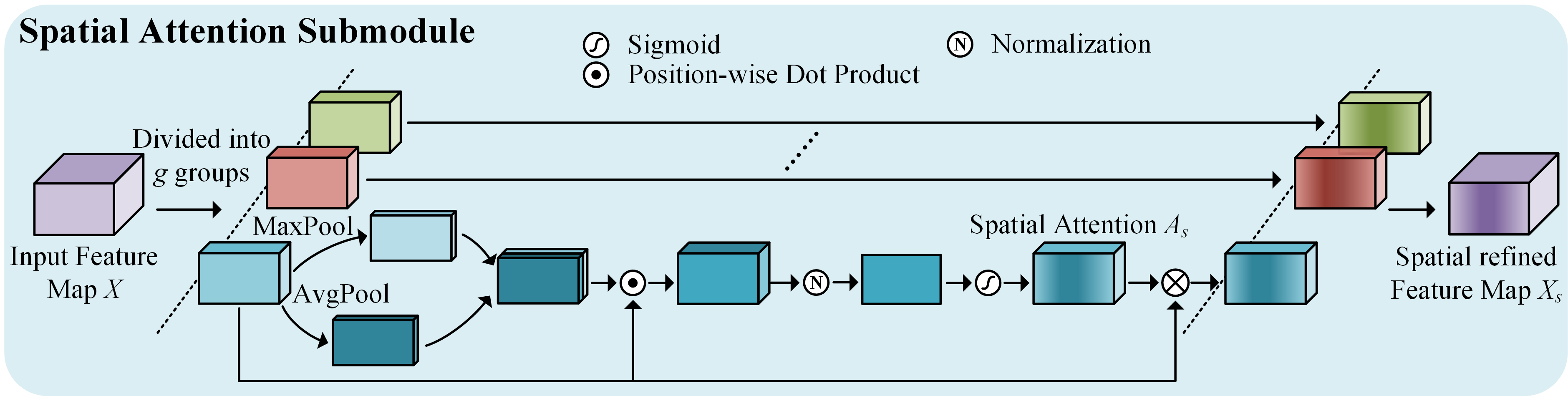}
\caption{Diagram of spatial attention submodule.}
\label{fig3}
\end{figure*}

The SGE attention module \cite{li2019spatial} has verified global avg-pooling is helpful to performance improvement. The reason is that avg-pooling calculates the mean value in the feature map region as the resulting pooled value, and can diminish the bias of estimated average value as well as improve the robustness of model. Based on this observation, we further introduce global max-pooling which calculates the maximum value in the feature map region as the resulting pooled value, as it is able to learn the edge information and texture structure of the feature map. We believe the combination of both these pooling methods can effectively aggregate the spatial information. Fig. \ref{fig3} depicts the computation process of spatial attention submodule. The detailed computation process is as follows.

(1) The input feature map $X \in \mathbb{R}^{C \times H \times W}$ is divided into groups along the direction of the channel. The grouping strategy is adopted due to two reasons. Firstly, it can well reduce the model computation overhead \cite{li2019spatial}. Secondly, as revealed in \cite{Sabour2017Capsule}, the learning process of DNNs can be regarded as a process of gradually capturing specific semantic responses, such as the bird's beak in the bird recognition task \cite{Fu2017Bird}. In this group space, ideally, we can obtain features with strong responses at the beak position, while other positions become zero vectors as no activation exists. However, it is generally difficult for DNN to obtain uniformly distributed feature responses, due to the existence of unavoidable noise and similar patterns \cite{li2019spatial}. Therefore, the overall information of the whole group space can be used to further enhance semantic feature learning at key regions.

Firstly, In the spatial dimension, each position of the group can be represented by the following vectors:

\begin{equation}
\label{eq3}
    \textbf{\emph{p}} = \{\textbf{P}_{1}, \ldots, \textbf{P}_{n}\},\textbf{P}_i \in \mathbb{R}^{\frac{C}{g}},n = H \times W,
\end{equation}

\noindent where $g$ is a pre-defined hyper-parameter denoting the number of groups, and $\textbf{P}_i$ denotes the local statistical feature. The similarity between the global statistical feature and the local one at each position can generate the spatial attention map. The global statistical feature can be obtained through avg-pooling and max-pooling:

\begin{equation}
\label{eq4}
    \textbf{f} = cat(AvgPool(\textbf{\emph{p}}),MaxPool(\textbf{\emph{p}})),
\end{equation}

\noindent where $cat$ denotes the concatenate operation.

(2) For each position $i \in \{1, \ldots, n\}$, the similarity between the global and local statistical features can be obtained by the simple dot product:

\begin{equation}
\label{eq5}
    W_i = \textbf{f} \cdot \textbf{P}_i = \left\|\textbf{f}\right\| \times \left\|\textbf{P}_i\right\| \times \cos{\theta_i},
\end{equation}

\noindent where $\theta_i$ is the angle between $\textbf{f}$ and $\textbf{P}_i$ \cite{li2019spatial}. To avoid bias among the coefficients of various samples, we then adopt normalization \cite{ioffe2015batch,qiao2019weight,wu2018group} to $W_i$:

\begin{equation}
\label{eq7}
	{{\hat N}_i} = \frac{{{W_i} - {\mu _w}}}{{{\sigma _w} + \varepsilon }}\\,
\end{equation}

\begin{equation}
{\mu _w} = \frac{1}{n}\sum\limits_j^n {{W_j}} \\,
\end{equation}

\begin{equation}
\sigma _w^2 = \frac{1}{n}\sum\limits_j^n {{{({W_j} - {\mu _w})}^2}},
\end{equation}
where $\varepsilon$ (e.g., 1e-5) is a constant added for numerical stability.

(3) The sigmoid function is used to calculate the final spatial attention map as follows:

\begin{equation}
\label{eq7}
    A_s = Sigmoid(N_i).
\end{equation}

\subsubsection{Channel Attention Submodule}

Different from spatial attention, the channel attention submodule focuses on ‘what’ are the informative parts and pays differentiated attention to different channels of feature map. For a feature map $X \in \mathbb{R}^{C \times H \times W}$, channel attention map $A_c \in \mathbb{R}^{C \times 1 \times 1}$ is corresponding to a $C \times 1 \times 1$ matrix, in which each position denotes a weight corresponding to a channel of original feature map.

The previously designed attention modules, e.g., SENet and CBAM, use two Fully-Connected (FC) layers to process channel information. There exist three drawbacks for such a design. Firstly, it limits the total number of attention modules that can be inserted into the original model \cite{yang2020gated}. Secondly, it becomes difficult to analyze the relationship between different layers of channels due to the complexity of parameters in FC layers. Actually, capturing the dependencies between all channels is inefficient and unnecessary. Thirdly, dimensionality reduction has to be involved to control model complexity, which has side effects on channel attention prediction \cite{wang2020eca}.

\begin{figure}[htb]
\centering
\includegraphics[scale=1.2]{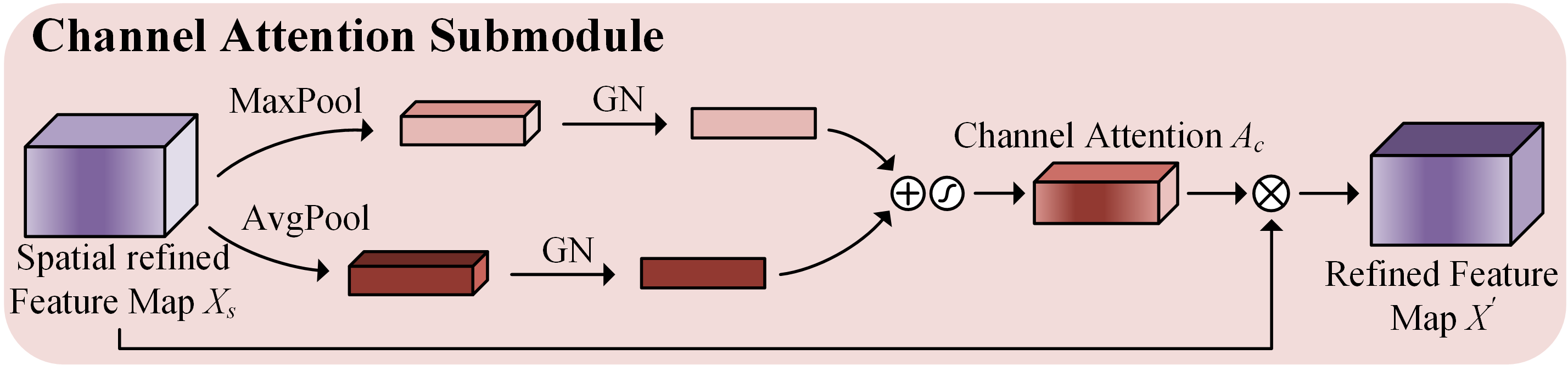}
\caption{Diagram of channel attention submodule.}
\label{fig4}
\end{figure}

To address these problems, we propose to use normalization to model the relationship of channels. Compared to the FC layers used in SENet and CBAM, Batch Normalization (BN) can generate competition relationship between channels, using much fewer resource cost while providing more stable training performance \cite{yang2020gated}. In this work, we choose Group Normalization (GN) \cite{wu2018group}, as a simple alternative of BN, to replace the design with two FC layers. In GN, the channels are divided into groups, and the mean and variance for normalization are computed within each group. As the computation of GN is independent of batch sizes, it can outperform BN and other normalization methods. Fig. \ref{fig4} illustrates the computation process of channel attention submodule. The detailed computation process is as follows.

(1) Due to similar aforementioned reasons, both avg-pooling and max-pooling are adopted in the channel attention submodule to aggregate the feature map in each channel, so as to generate two different spatial context descriptors as follows:

\begin{equation}
\label{eq8}
    A = AvgPool(X_s),
\end{equation}

\begin{equation}
\label{eq9}
    M = MaxPool(X_s),
\end{equation}

\noindent where $A$ denotes the average-pooled features, and $M$ denotes the max-pooled features.

(2) Then, $A$ and $M$ are normalized respectively by GN. Take $A$ as an illustration. Given $A$, GN performs the following computation:
\begin{equation}
\label{eq10}
    \hat{A_i} = \frac{1}{\sigma_i}(A_i - \mu_i),
\end{equation}

\noindent where $i$ denotes the index. For 2D images, $i=(i_N,i_C,i_H,i_W)$ is a 4D vector indexing the features in the order of $(N,C,H,W)$, where $N$ is the batch axis. Meanwhile, $\mu$ and $\sigma$ in (\ref{eq10}) are the mean and the standard deviation (std), respectively. They can be calculated as:

\begin{equation}
\label{eq11}
    \mu_i = \frac{1}{m}\sum_{K \in \Gamma_i}A_K,
\end{equation}

\begin{equation}
 \label{eq12}
    \sigma_i = \sqrt[]{\frac{1}{m}\sum_{K \in \Gamma_i}(A_K - \mu_i)^{2} + \epsilon},
\end{equation}

\noindent where $\epsilon$ is a small constant, $\Gamma_i$ is the set of pixels in which the mean and the std are computed, and $m$ is the size of this set. Actually, the primary difference between the various feature normalization methods is the different definition of $\Gamma_i$ \cite{ioffe2015batch,ba2016layer,wu2018group}. In GN, this set is defined as:

\begin{equation}
\label{eq13}
    \Gamma_i = \{K|K_N = i_N,\lfloor\frac{K_C}{C / G}\rfloor = \lfloor\frac{i_C}{C / G}\rfloor \},
\end{equation}

\noindent where $G$ is a pre-defined hyper-parameter denoting the number of groups, and $C/G$ is the number of channels in each group. The normalization operation for $M$ follows a similar procedure, and therefore is omitted for simplicity.

(3) We merge the normalized output feature vectors using element-wise summation, and generate the 3D channel attention map $A_c$ via a sigmoid function. The final channel attention map is computed as:

\begin{equation}
    \begin{aligned}
    A_c &= Sigmoid(GN(MaxPool(X_s)) + GN(AvgPool(X_s)))\\
        &= Sigmoid(GN(M)+GN(A)). \label{eq:no14}\\
    \end{aligned}
\end{equation}

\subsubsection{How to Combine Two Attention Submodules}

The spatial and channel attention submodules can complement each other very well on image classification, as they focus on `where' and `what' is meaningful respectively. How to arrange them has a dramatic impact on the final performance, and should be taken into account when design. Actually, they can be combined in a sequential manner (e.g., CBAM), or in a parallel manner (e.g., BAM). By experiments, we found that the sequential arrangement with the spatial-first order achieves the best result. That's why we name our module as ``SCA''. Detailed experimental results will be presented in the next section.

\subsection{Channel Pruning Guided by SCA}

Attention mechanism can explicitly describes the importance relationship between channels in the same layer, and constantly adjusts the parameters of fully-connected layers in the process of back-propagation \cite{xu2015show,woo2018cbam}. By inserting the attention module, the network is capable of showing the trend of enhancing or suppressing a portion of channels gradually. We propose to use the channel scale that is a statistical quantity found by element-wise averaging of the weight in the channel attention map over all training data, as a criterion for measuring channel importance. Note that for different input data (i.e.,images), the attention module will output different weight outputs for the same channel \cite{gao2018dynamic}, as illustrated by the experimental results in Fig. \ref{fig5}. As a result, the channel importance in CPSCA is measured in a statistical fashion for fairness and precision \cite{yamamoto2018pcas}. The channel scale for a given channel $j$ is given as follows:

\begin{figure}[htb]
\centering
\includegraphics[scale=0.4]{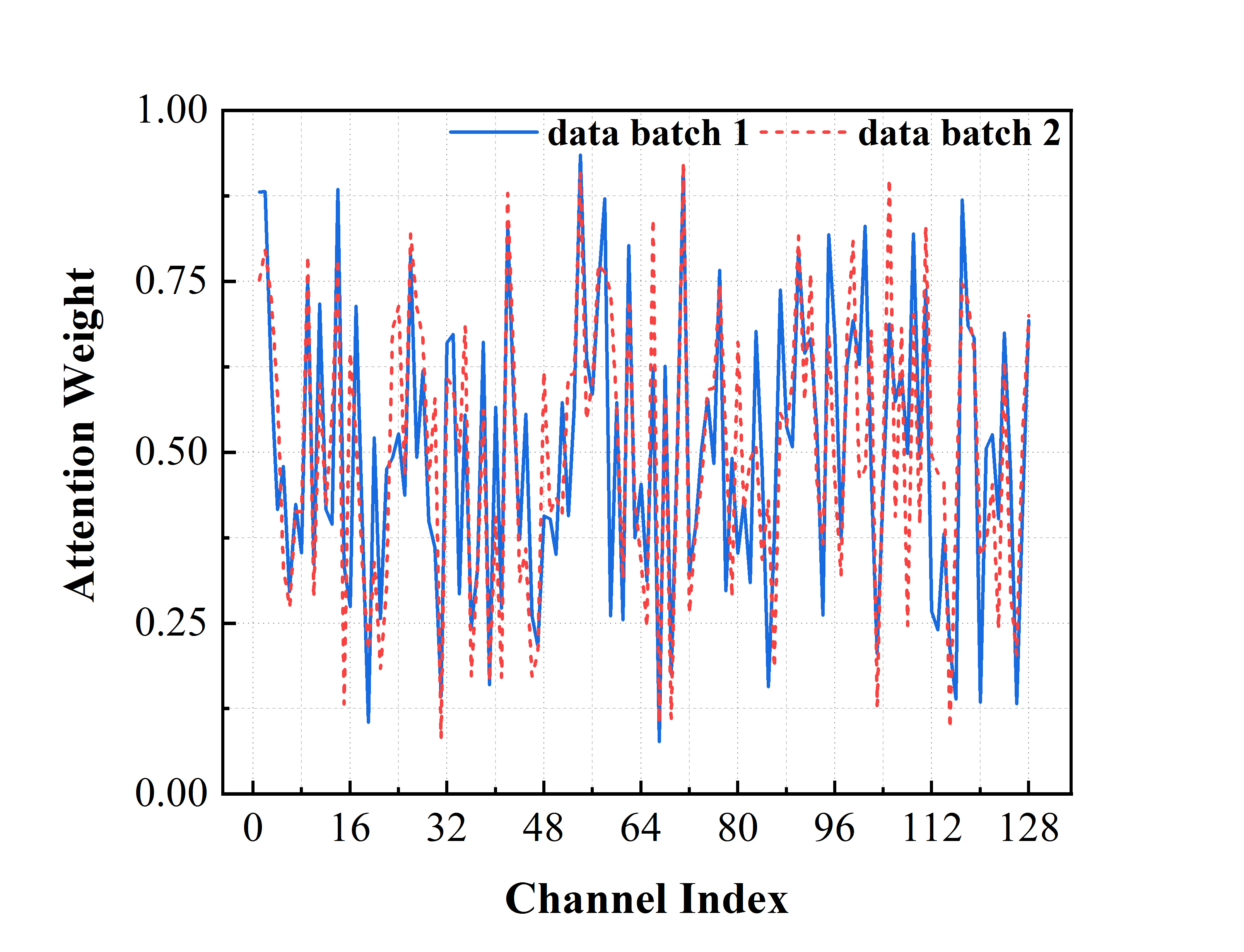}
\caption{An illustration of comparison results for the attention weights generated from different input data.}
\label{fig5}
\end{figure}

\begin{equation}
\label{eq15}
        w_j = \frac{1}{|D|}\sum_{d \in D}S_{j}(A_{c}^d), j \in \{1, \ldots, C_l\}, l \in \{1, \ldots, L\}.
\end{equation}

\noindent where $D$ denotes the set of training data, $A_{c}^d$ denotes the channel attention map for input data $d$, $L$ denotes the number of layers, and $S_j$ denotes the function that extracts the weight value for channel $j$ from the $C \times 1 \times 1$ channel attention matrix.

The overall channel pruning process in CPSCA is summarized in Algorithm \ref{alg1}.

\begin{algorithm}[htb]
\caption{\textbf{CPSCA Algorithm}}
\label{alg1}
\begin{algorithmic}[1] 

\REQUIRE ~~\\ 
$M$: the original DNN model \\
$L$: the number of layers in $M$ \\
$D$: the set of training data \\
$w_j$: the weight value for channel $j$ output by SCA \\
$\{p_l\%|l\in \{1, ..., L\}\}$: the pruning ratios for all $L$ layers

\ENSURE ~~\\ 
 $M^{'}$: the pruned DNN model\\

\STATE \textbf{Procedure}
\STATE Insert SCA modules into model $M$

\FOR{each batch $d \in D$}
    \STATE Train model $M$ with $d$
    \FOR{each layer $l \in \{1,...,L\}$}
        \FOR{each channel $j \in \{1,...,C_l\}$}
            \STATE Calculate the attention weight $S_{j}(A_{c}^d)$ for $j$
        \ENDFOR
    \ENDFOR
\ENDFOR

\FOR{each layer $l \in \{1,...,L\}$}
    \FOR{each channel $j \in \{1,...,C_l\}$}
        \STATE $w_j = \frac{1}{|D|}\sum_{d \in D}S_{j}(A_{c}^d)$
    \ENDFOR
    \STATE Sort all the $C_l$ channels by $w_j$, $j \in \{1, \ldots, C_l\}$
\ENDFOR

\STATE Remove all inserted SCA modules

\STATE Prune all layer in model $M$, i.e., removing a proportion $p_l\%$ of channels with the least scale values from layer $l$. Note that the filters in the next convolutional layer corresponding to the pruned channels are also removed

\STATE Finetune the model for accuracy recovery, and obtain $M'$

\RETURN $M^{'}$  
\end{algorithmic}
\end{algorithm}

\section{Performance Analysis}

In this section, we evaluate SCA and CPSCA on the standard benchmarks CIFAR-10 and CIFAR-100 for image classification. The CIFAR-10 dataset consists of 60000 32$\times$32 color images in 10 classes, with 6000 images per class. There are 50000 training images and 10000 test images. The CIFAR-100 dataset is just like the CIFAR-10, except it has 100 classes containing 600 images each. There are 500 training images and 100 testing images per class. We perform channel pruning on VGGs and ResNets. Firstly, we perform extensive ablation experiments to fully evaluate the effectiveness of the final SCA module. Next, we demonstrate the applicability and performance of SCA across typical architectures and different datasets, as compared to the previously published attention modules. At last, we show that with the guidance of SCA, our CPSCA approach outperforms the state-of-the-art pruning methods. Moreover, we deploy the pruned model onto NVIDIA Jetson Nano \cite{Hadidi2019} to demonstrate the execution speedup of DNN models on the real edge platform. All the evaluation models are trained on an 8 Nvidia Titan-X GPUs server using Pytorch.

\subsection{Ablation studies}

This subsection shows the effectiveness of our design choice for the attention module. We first carefully search for the optimal structure design of the spatial attention, and then the channel attention. Then, we find out the best way of arrange the spatial and channel attention submodules through comparison experiments. We will explain the experiments for this module design process in detail as follows.

\subsubsection{Spatial Attention Submodule}

In this part of experiments, we only place the spatial attention submodule in SCA.

\textbf{Groups:} In the spatial attention submodule, we first investigate the impact of the number of groups, $g$, which denotes the number of different semantic sub-features. When the number of channels in the same convolutional layer is fixed, too few groups are not conducive to semantic diversity; On the contrary, too many groups will make feature representation for each semantic response limited. From Fig. \ref{fig6a}, we can observe that $g=64$ tends to produce better prediction results than the others. It’s a moderate value that can balance semantic diversity and representation ability of each semantic to optimize inference performance.

\textbf{Pooling:} In order to verify the effectiveness of the joint using of both poolings in spatial attention module, we compare three different pooling strategies: avg-pooling, max-pooling and joint use of both poolings. The results in Fig. \ref{fig6b} confirm that joint using both avg-pooling and max-pooling significantly enhances representation power of DNN models, resulting in higher accuracy than using each independently. That's because avg-pooling extracts features smoothly while max-pooling focuses only on the most salient features. It is better to use them simultaneously to make them compensate with each other.

As a brief summary, we use both poolings in our spatial attention submodule, and the number of groups $g = 64$ in the following experiments.

\subsubsection{Channel Attention Submodule}

After the spatial-wise refined features are given, we can further explore how to effectively compute the channel attention. In this part of experiments, we place the channel attention submodule just after the previously designed spatial attention submodule in SCA, since our ultimate goal is to combine them together.

\begin{figure*}[tb] 
    \centering  
    \subfigure{ 
        \includegraphics[width=0.23\linewidth]{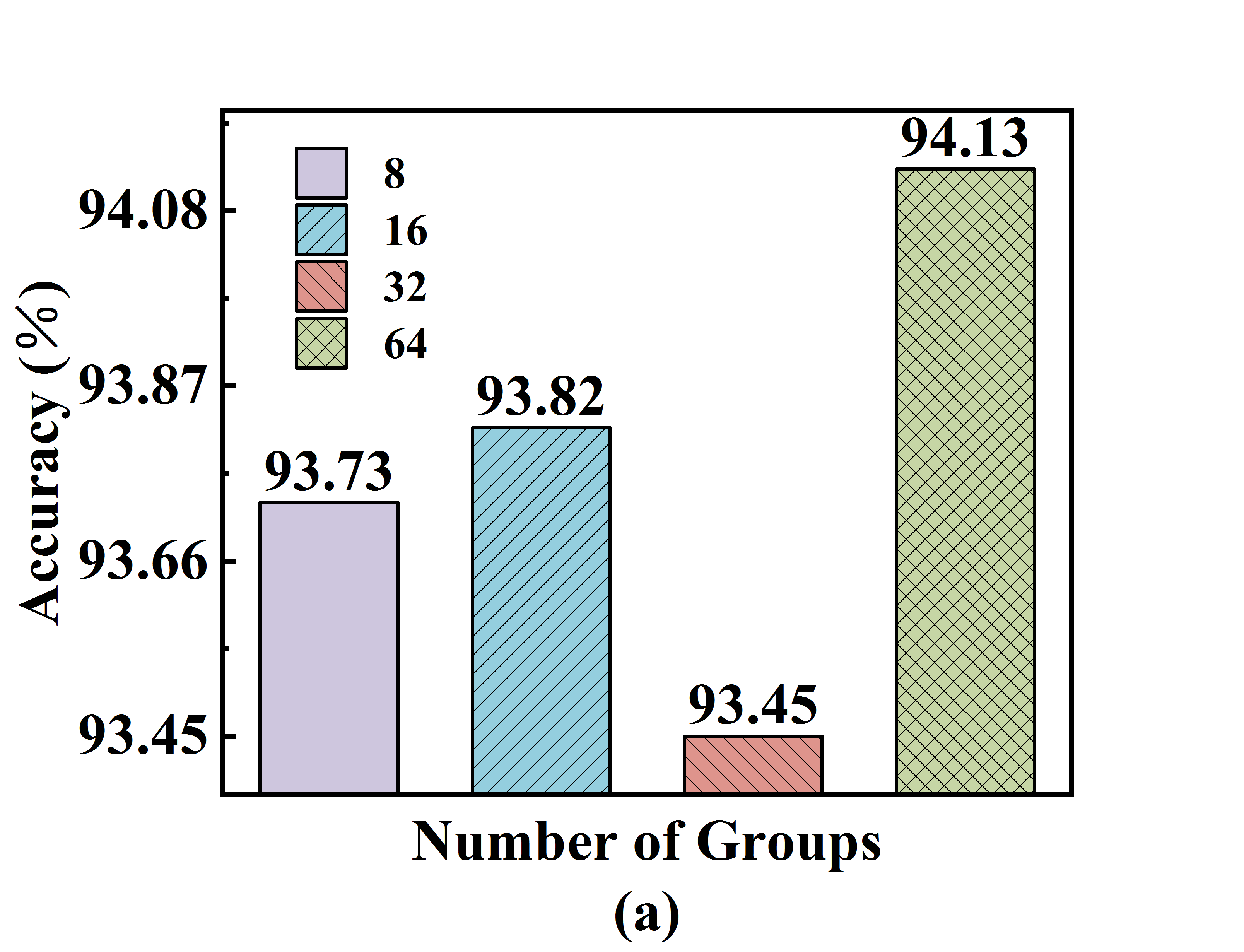}\label{fig6a}}
    \subfigure{
        \includegraphics[width=0.23\linewidth]{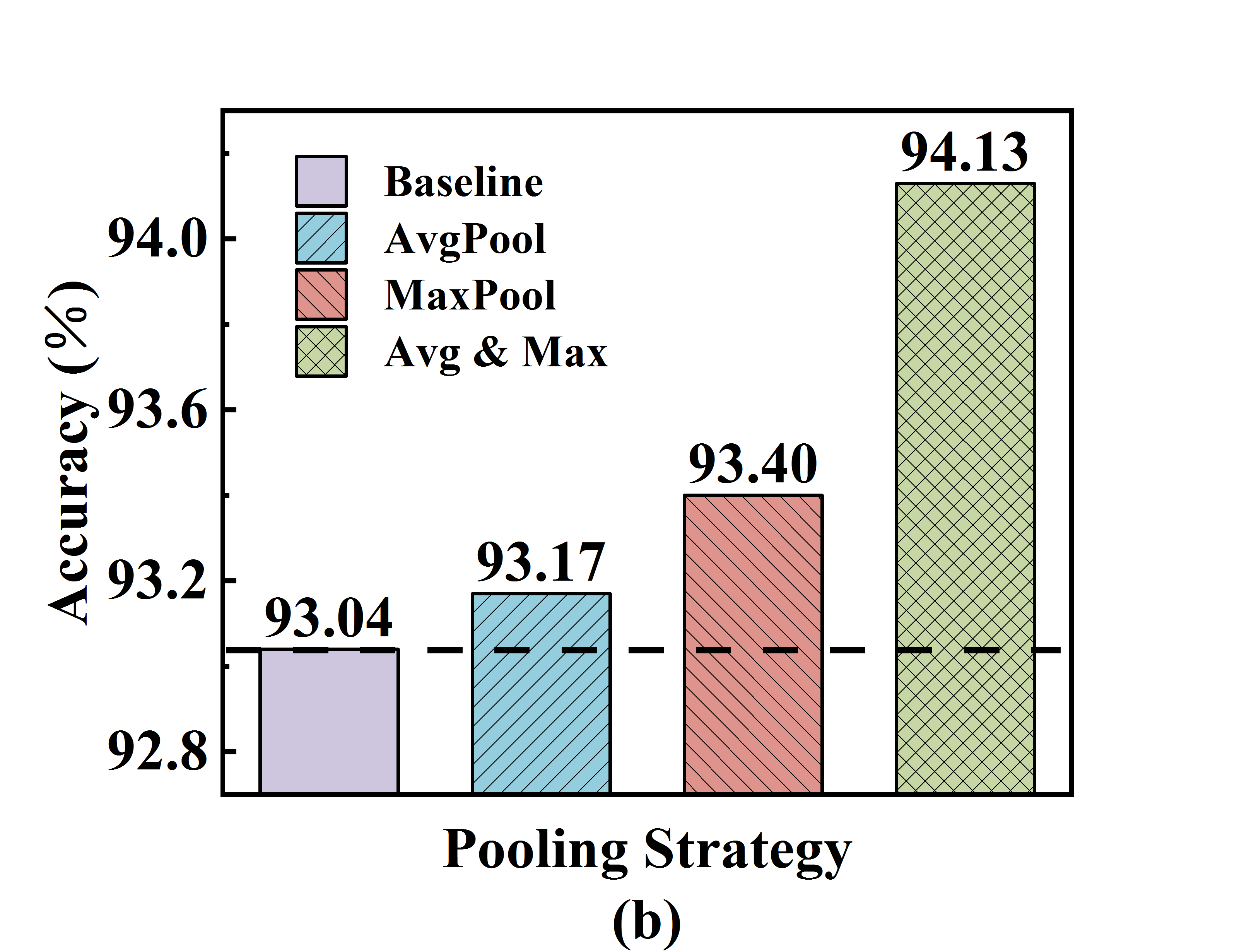}\label{fig6b}}
    \subfigure{
        \includegraphics[width=0.23\linewidth]{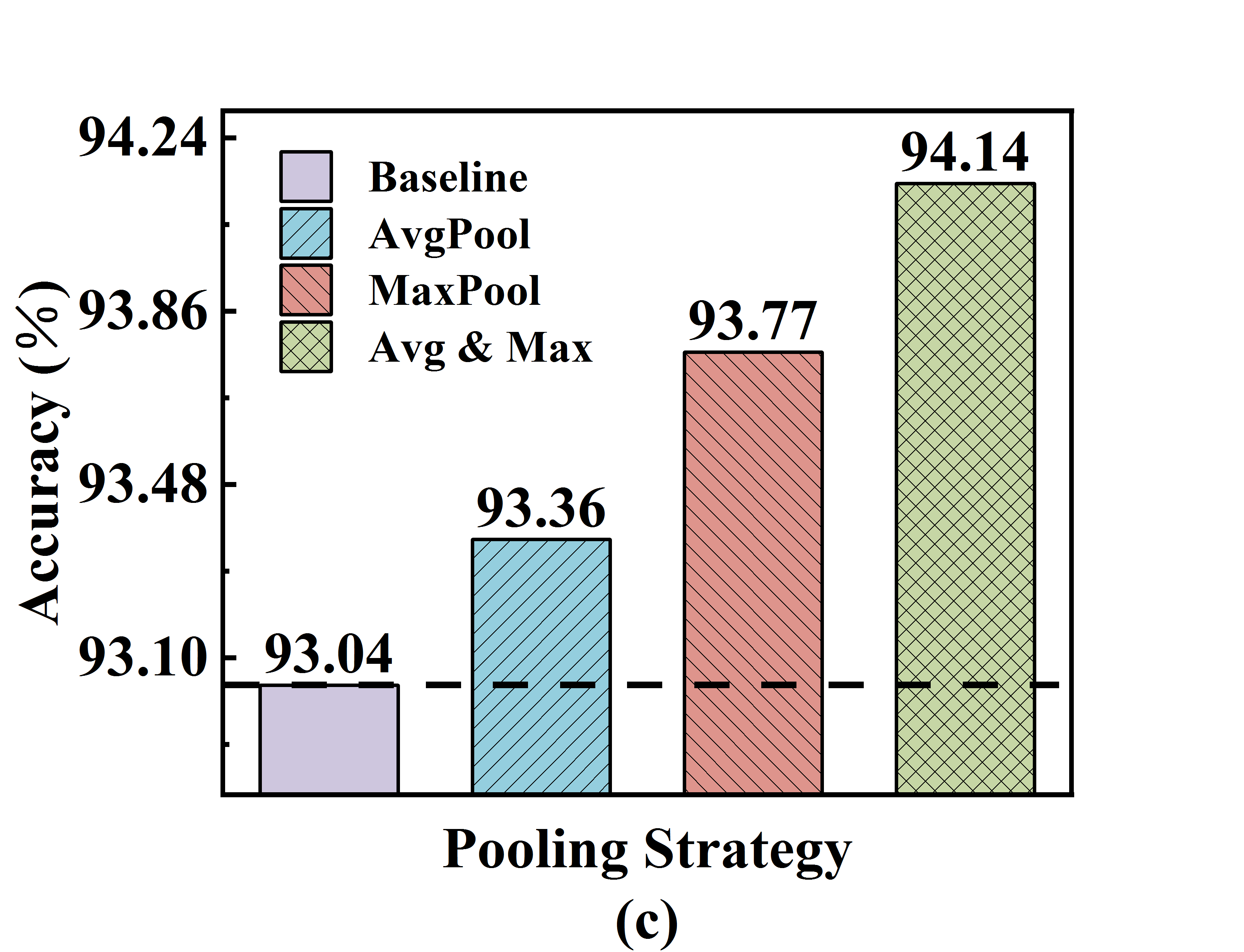}\label{fig6c}}
    \subfigure{
        \includegraphics[width=0.23\linewidth]{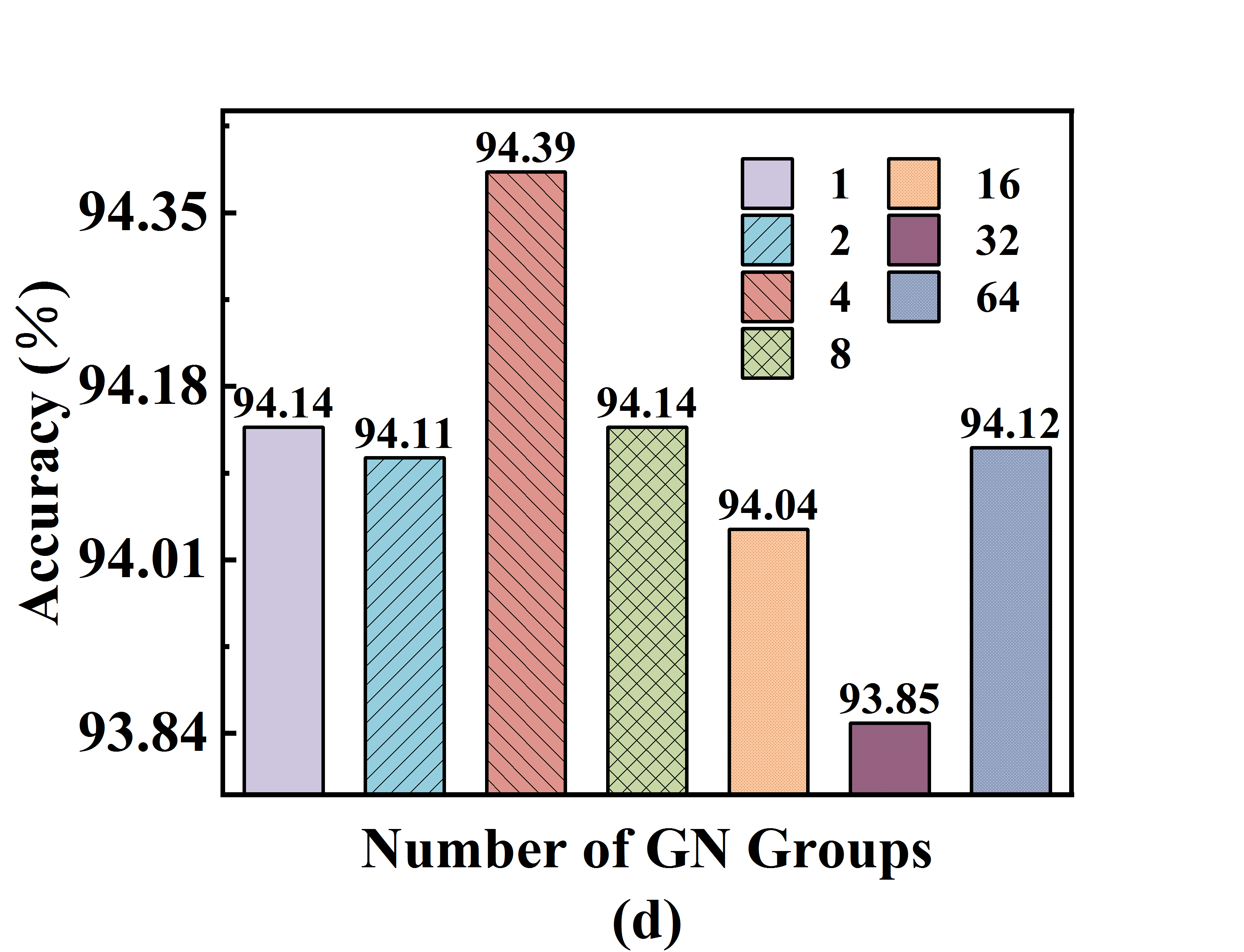}\label{fig6d}}
    \caption{(a) Number of groups $g$ vs. performance. Evaluation uses VGG16 on CIFAR-10 dataset with both poolings. (b) Pooling strategies vs. performance. Evaluation uses VGG16 on CIFAR-10 dataset with $g = 64$. (c) Pooling strategies vs. performance. Evaluation uses VGG16 on CIFAR-10 dataset with $G = 1$. (d) Number of GN groups $G$ vs. performance. Evaluation uses VGG16 on CIFAR-10 dataset with both poolings.}
    \label{level}
\end{figure*}

\textbf{Pooling:} Same as that in spatial attention, both avg-pooling and max-pooling are adopted in the channel attention module. Fig. \ref{fig6c} shows the experimental results with the typical pooling strategies. Similar to the results in spatial attention, all the three pooling strategies outperform the baseline which does not use pooling, and the best accuracy result can be achieved by joint using both poolings.

\textbf{GN groups:} In the channel attention submodule, we apply Group Normalization (GN) to both avg-pooled and max-pooled features simultaneously. In the experiments, the setting of group division is the same as that in \cite{wu2018group}. From Fig. \ref{fig6d}, GN performs well for all values of $G$, and the inference accuracies increase observably by 0.81\%-1.35\% as compared to the baseline. Meanwhile, the results indicate that setting $G = 4$ achieves the best performance among all the options.

As a brief summary, we also use both poolings in our channel attention submodule, and the number of GN groups $G = 4$ in the following experiments.

\subsubsection{Arrangement of the spatial and channel attention}

After determining the suitable settings for each of the two submodules, we compare five different ways of arranging these submodules in SCA (i.e., only channel, only spatial, sequential spatial and channel, sequential channel and spatial, and parallel use of both submodules), to investigate how the existence and the order of submodules could affect the overall performance. From the results summarized in Table \ref{tb1}, all the arranging patterns outperform the baseline, indicating the advantage of applying attention mechanisms. We can also observe that the sequential arrangement surpasses the parallel arrangement, and the spatial-first order achieves better results than the channel-first order. According to the comparison results, we choose to arrange the spatial and channel attention submodules sequentially in SCA.
\begin{table}[H]
\centering
\renewcommand\tabcolsep{2.21pt} 
\caption{Comparison of different combining methods using VGG16 on CIFAR-10 dataset.}
\label{tb1}
\small
\begin{tabular}{@{}lccc@{}}
\toprule
\textbf{Description}                    & \textbf{Params} & \textbf{GFLOPs} & \textbf{Acc(\%)} \\ \midrule
VGG16 (baseline)                                  & 16.87M              & 0.63163         & 93.04 \\
VGG16 + Spatial                         & 16.87M              & 0.63163         & 94.13           \\
VGG16 + Channel                         & 16.88M              & 0.63163         & 93.82           \\
\textbf{VGG16 + Spatial + Channel (ours)}    & 16.88M              & 0.63163         & \textbf{94.39}  \\
VGG16 + Channel + Spatial               & 16.88M              & 0.63163         & 93.61           \\
VGG16 + Channel \& Spatial in parallel & 16.88M              & 0.63163         & 93.57           \\ \bottomrule
\end{tabular}
\end{table}

\subsection{Comparisons with State-of-the-art Attention Modules}

We perform image classification experiments to evaluate our SCA module in two popular backbone architectures, i.e., VGG and ResNet. In VGG networks the SCA module is placed at every convolutional block, and in ResNet networks the SCA module is placed on the convolution outputs in each ResBlock. We compare SCA against several state-of-the-art attention modules, including SENet, CBAM and SGE, based on CIFAR-10 and CIFAR-100 datasets. The evaluation metrics consist of two aspects: efficiency (i.e., parameter size, and computation cost) and effectiveness (i.e., Top-1 accuracy and accuracy improvement). The comparison results are summarized in Table \ref{tb2} and Table \ref{tb3}.

\begin{table}[H]
\centering
\renewcommand\tabcolsep{6.4pt} 
\caption{Comparison of different attention modules on CIFAR-10 dataset.}
\label{tb2}
\small
\begin{tabular}{clcccc}
\toprule
\multicolumn{1}{c}{\textbf{Model}}     & \textbf{Variant}      & \textbf{Params}    & \textbf{GFLOPs}    & \multicolumn{2}{c}{\textbf{Acc(\%)}}   \\ \midrule
 & baseline       & 16.87M    & 0.63163     & 93.04       & —   \\
 & + SENet  & 17.10M    & 0.63209     & 93.50      & $\uparrow$0.46    \\
 & + CBAM       & 17.10M    & 0.63310     & 93.19     & $\uparrow$0.15    \\
& + SGE    & 16.87M   & 0.63163   & 93.12   & $\uparrow$0.08\\
\multirow{-5}{*}{VGG16}                         & \textbf{+ SCA} & 16.88M & 0.63163 & \textbf{94.39} & \textbf{$\uparrow$1.35}   \\ \midrule
& baseline                & 22.18M                      & 0.80158                      & 92.87                                 & —                                                            \\
                                                & + SENet                                         & 22.49M                      & 0.80219                      & 93.33                                 & $\uparrow$0.46                                                        \\
                                                & + CBAM                                       & 22.49M                      & 0.80336                      & 93.59                                 & $\uparrow$0.72                                                        \\
                                                & + SGE                                        & 22.18M                     & 0.80158                      & 92.92                                 & $\uparrow$0.05                                                        \\
\multirow{-5}{*}{VGG19}                         & \textbf{+ SCA} & 22.19M & 0.80158 & \textbf{93.97}   & \textbf{$\uparrow$1.10}                          \\ \midrule
                                                & baseline                                     & 0.85M                       & 0.25257                      & 92.98                                 & —                                                            \\
                                                & + SENet                                         & 0.87M                       & 0.25260                      & 93.53                                 & $\uparrow$0.55                                                        \\
                                                & + CBAM                                       & 0.87M                       & 0.25742                      & 93.61                                 & $\uparrow$0.63                                                        \\
                                                & + SGE                                        & 0.85M                       & 0.25257                      & 93.78                                 & $\uparrow$0.80                                                        \\
\multirow{-5}{*}{ResNet56}                      & \textbf{+ SCA} & 0.86M  & 0.25257 & \textbf{93.86}   & \textbf{$\uparrow$0.88}                          \\ \midrule
\multicolumn{1}{c}{}                            & baseline                                     & 1.73M                       & 0.50893                      & 93.21                                 & —                                                            \\
\multicolumn{1}{c}{}                            & + SENet                                         & 1.76M                       & 0.50898                      & 93.53                                 & $\uparrow$0.32                                                        \\
\multicolumn{1}{c}{}                            & + CBAM                                       & 1.77M                       & 0.51861                      & 93.43                                 & $\uparrow$0.22                                                        \\
\multicolumn{1}{c}{}                            & + SGE                                        & 1.73M                       & 0.50893                      & 93.81                                 & $\uparrow$0.60                                                        \\
\multicolumn{1}{c}{\multirow{-5}{*}{ResNet110}} & \textbf{+ SCA} & 1.74M  & 0.50893 & \textbf{93.87}   & \textbf{$\uparrow$0.66}                          \\ \bottomrule
\end{tabular}
\end{table}

\begin{table}[htb]
\centering
\renewcommand\tabcolsep{7.4pt} 
\caption{Comparison of different attention modules on CIFAR-100 dataset.}
\label{tb3}
\small
\begin{tabular}{@{}clcccc@{}}
\toprule
\multicolumn{1}{c}{\textbf{Model}} & \multicolumn{1}{c}{\textbf{Variant}} & \textbf{Params}         & \textbf{GFLOPs}              & \multicolumn{2}{c}{\textbf{Acc(\%)}}                     \\ \midrule
                                   & baseline                                 & 17.24M                      & 0.63237                      & 73.14                      & —                        \\
                                   & + SENet                                     & 17.47M                      & 0.63282                      & 74.87                      & $\uparrow$1.73                      \\
                                   & + CBAM                                   & 17.47M                      & 0.63384                      & 75.14                      & $\uparrow$2.00                      \\
                                   & + SGE                                    & 17.24M                      & 0.63237                      & 73.25                      & $\uparrow$0.11                      \\
\multirow{-5}{*}{VGG16}            & \textbf{+ SCA}      & 17.25M & 0.63237 & \textbf{75.24} & \textbf{$\uparrow$2.10} \\ \midrule
                                   & baseline            & 22.55M                      & 0.80232                      & 73.02                      & —                        \\
                                   & + SENet                                     & 22.85M                      & 0.80292                      & 74.34                      & $\uparrow$1.32                      \\
                                   & + CBAM                                   & 22.86M                      & 0.80410                      & 74.12                      & $\uparrow$1.10                      \\
                                   & + SGE                                    & 22.55M                      & 0.80232                      & 73.62                      & $\uparrow$0.60                      \\
\multirow{-5}{*}{VGG19}            & \textbf{+ SCA}      & 22.56M & 0.80232 & \textbf{74.74} & \textbf{$\uparrow$1.72} \\ \midrule
                                   & baseline                                 & 0.86M                       & 0.25258                      & 75.78                      & —                          \\
                                   & + SENet                                     & 0.87M                       & 0.25261                      & 76.12                      & $\uparrow$0.34                      \\
                                   & + CBAM                                   & 0.88M                       & 0.25743                      & 76.09                      & $\uparrow$0.31                      \\
                                   & + SGE                                    & 0.86M                       & 0.25258                      & 76.28                      & $\uparrow$0.50                      \\
\multirow{-5}{*}{ResNet56}         & \textbf{+ SCA}      & 0.86M  & 0.25258 & \textbf{76.73} & \textbf{$\uparrow$0.95} \\ \midrule
                                   & baseline                                 & 1.73M                       & 0.50894                      & 75.87                      & —                          \\
                                   & + SENet                                     & 1.76M                       & 0.50899                      & 76.06                      & $\uparrow$0.04                      \\
                                   & + CBAM                                   & 1.77M                       & 0.51862                      & 76.18                      & $\uparrow$0.31                      \\
                                   & + SGE                                    & 1.74M                       & 0.50894                      & 76.15                      & $\uparrow$0.28                      \\
\multirow{-5}{*}{ResNet110}        & \textbf{+ SCA}      & 1.75M  & 0.50894 & \textbf{76.27} & \textbf{$\uparrow$0.40} \\ \bottomrule
\end{tabular}
\end{table}

\begin{table*}[!htb]
\centering
\renewcommand\tabcolsep{1.7pt} 
\caption{Comparison of different pruning approaches using VGG16 on CIFAR-10 and CIFAR-100 dataset.}
\label{tb4}
\small
\begin{tabular}{l|ccccccc|ccccccc}
\hline
    \multicolumn{1}{c|}{\multirow{2}{*}{\textbf{Method}}} & \multicolumn{7}{c|}{\textbf{CIFAR-10}}  & \multicolumn{7}{c}{\textbf{CIFAR-100}} \\ \cline{2-15}
    \multicolumn{1}{c|}{}  & \textbf{Params} & \textbf{Pruned}  & \textbf{GFLOPs}   & \textbf{Pruned} & \textbf{Time(ms)}   & \multicolumn{2}{c|}{\textbf{Acc(\%)}} & \textbf{Params}     & \textbf{Pruned}   & \textbf{GFLOPs}   & \textbf{Pruned} & \textbf{Time(ms)}  & \multicolumn{2}{c}{\textbf{Acc(\%)}} \\ \hline
    \textbf{baseline}                                     & 16.87M                  & —                        & 0.63163                  & —                      & 12.3  & 93.04             & —                & 17.24M                  & —                        & 0.63237                  & —                       & 13.5 & 73.14            & —                \\ \hline
$\ell_{1}$-norm                                      & \multirow{4}{*}{11.43M} & \multirow{4}{*}{32.25\%} & \multirow{4}{*}{0.47499} & \multirow{4}{*}{24.75\%}  & \multirow{4}{*}{10.2} & 93.25             & $\uparrow$0.21            & \multirow{4}{*}{11.47M} & \multirow{4}{*}{33.47\%} & \multirow{4}{*}{0.47509} & \multirow{4}{*}{24.87\%} & \multirow{4}{*}{9.6} & 72.97            & $\downarrow$0.17            \\
Slimming                           &                         &                          &                          &                          & &93.35             & $\uparrow$0.31            &                         &                          &                          &                          & &73.34            & $\uparrow$0.20            \\
CPSE                                     &                         &                          &                          &                          & &93.43             & $\uparrow$0.39            &                         &                          &                          &                          & &73.36            & $\uparrow$0.22            \\
\textbf{CPSCA}                 &                         &                          &                          &                         & & \textbf{93.45}    & \textbf{$\uparrow$0.41}   &                         &                          &                          &                       &   & \textbf{73.77}   & \textbf{$\uparrow$0.63}   \\ \hline
$\ell_{1}$-norm                                      & \multirow{4}{*}{8.52M}  & \multirow{4}{*}{49.50\%} & \multirow{4}{*}{0.21988} & \multirow{4}{*}{65.05\%}  &\multirow{4}{*}{4.5} & 92.98             & $\downarrow$0.06            & \multirow{4}{*}{8.57M}  & \multirow{4}{*}{50.29\%} & \multirow{4}{*}{0.21997} & \multirow{4}{*}{65.21\%}  &\multirow{4}{*}{4.2} & 71.76            & $\downarrow$1.38            \\
Slimming                           &                         &                          &                          &                         & & 93.01             & $\downarrow$0.03            &                         &                          &                          &                        &  & 71.95            & $\downarrow$1.19            \\
CPSE                                     &                         &                          &                          &                         & & 92.09             & $\downarrow$0.95            &                         &                          &                          &                        &  & 72.13            & $\downarrow$1.01            \\
\textbf{CPSCA}                   &                         &                          &                          &                         & & \textbf{93.10}    & \textbf{$\uparrow$0.06}   &                         &                          &                          &                         & & \textbf{72.89}   & \textbf{$\downarrow$0.25}   \\ \hline
$\ell_{1}$-norm                                      & \multirow{4}{*}{3.38M}  & \multirow{4}{*}{79.96\%} & \multirow{4}{*}{0.14407} & \multirow{4}{*}{77.02\%} &\multirow{4}{*}{3.0} & 92.13             & $\downarrow$0.91            & \multirow{4}{*}{3.89M}  & \multirow{4}{*}{77.44\%} & \multirow{4}{*}{0.18677} & \multirow{4}{*}{70.47\%} &\multirow{4}{*}{3.9} & 70.19            & $\downarrow$2.95            \\
Slimming                           &                         &                          &                          &                         & & 92.78             & $\downarrow$0.26            &                         &                          &                          &                         & & 71.33            & $\downarrow$1.81            \\
CPSE                                     &                         &                          &                          &                       &   & 92.89             & $\downarrow$0.15            &                         &                          &                          &                          & & 71.09            & $\downarrow$2.05            \\
\textbf{CPSCA}                 &                         &                          &                          &                          & & \textbf{92.96}    & \textbf{$\downarrow$0.08}   &                         &                          &                          &                         & & \textbf{72.03}   & \textbf{$\downarrow$1.11}   \\ \hline
$\ell_{1}$-norm                                      & \multirow{4}{*}{0.72M}  & \multirow{4}{*}{95.73\%} & \multirow{4}{*}{0.09918} & \multirow{4}{*}{84.12\%} &\multirow{4}{*}{2.1} & 88.90             & $\downarrow$4.14            & \multirow{4}{*}{0.77M}  & \multirow{4}{*}{95.53\%} & \multirow{4}{*}{0.09927} & \multirow{4}{*}{84.30\%} &\multirow{4}{*}{2.1} & 69.87            & $\downarrow$3.27            \\
Slimming                           &                         &                          &                          &                          && 89.79             & $\downarrow$3.25            &                         &                          &                          &                          & & 70.25            & $\downarrow$2.89            \\
CPSE                                     &                         &                          &                          &                        &  & 90.11             & $\downarrow$2.93            &                         &                          &                          &                          & & 70.87            & $\downarrow$2.27            \\
\textbf{CPSCA}                 &                         &                          &                          &                        &  & \textbf{91.39}    & \textbf{$\downarrow$1.65}   &                         &                          &                          &                          & & \textbf{71.67}   & \textbf{$\downarrow$1.47}   \\ \hline
\end{tabular}
\end{table*}

\begin{table*}[!htb]
\centering
\renewcommand\tabcolsep{1.7pt} 
\caption{Comparison of different pruning approaches using ResNet56 on CIFAR-10 and CIFAR-100 dataset.}
\label{tb5}
\small
\begin{tabular}{l|ccccccc|ccccccc}
\hline
    \multicolumn{1}{c|}{\multirow{2}{*}{\textbf{Method}}} & \multicolumn{7}{c|}{\textbf{CIFAR-10}}  & \multicolumn{7}{c}{\textbf{CIFAR-100}} \\ \cline{2-15}
    \multicolumn{1}{c|}{}                                 & \textbf{Params}         & \textbf{Pruned}          & \textbf{GFLOPs}          & \textbf{Pruned}     & \textbf{Time(ms)}       & \multicolumn{2}{c|}{\textbf{Acc(\%)}}  & \textbf{Params}         & \textbf{Pruned}          & \textbf{GFLOPs}          & \textbf{Pruned}       & \textbf{Time(ms)}        & \multicolumn{2}{c}{\textbf{Acc(\%)}}  \\ \hline
\textbf{baseline}                                              & 0.86M                  & —                        & 0.25257                  & —                  &30.6      & 92.98          & —              & 0.86M                  & —                        & 0.25258                  & —                    &30.9    & 75.78          & —              \\ \hline
$\ell_{1}$-norm                                               & \multirow{4}{*}{0.45M} & \multirow{4}{*}{47.67\%} & \multirow{4}{*}{0.16983} & \multirow{4}{*}{32.76\%} & \multirow{4}{*}{20.7} & 92.54          & $\downarrow$0.44          & \multirow{4}{*}{0.45M} & \multirow{4}{*}{47.67\%} & \multirow{4}{*}{0.16984} & \multirow{4}{*}{32.76\%} & \multirow{4}{*}{21.0} & 74.89          & $\downarrow$0.89          \\
Slimming                                    &                        &                          &                          &                        &  & 92.26          & $\downarrow$0.72          &                        &                          &                          &                         & & 75.60          & $\downarrow$0.18          \\
CPSE                                              &                        &                          &                          &                          & &92.55          & $\downarrow$0.43          &                        &                          &                          &                         & & 75.55          & $\downarrow$0.23          \\
\textbf{CPSCA}                                          &                        &                          &                          &                        &  & \textbf{93.06} & \textbf{$\uparrow$0.08} &                        &                          &                          &                         & & \textbf{75.87} & \textbf{$\uparrow$0.09} \\ \hline
$\ell_{1}$-norm                                               & \multirow{4}{*}{0.17M} & \multirow{4}{*}{80.23\%} & \multirow{4}{*}{0.06476} & \multirow{4}{*}{74.36\%} & \multirow{4}{*}{8.1} & 90.48          & $\downarrow$2.50          & \multirow{4}{*}{0.18M} & \multirow{4}{*}{79.07\%} & \multirow{4}{*}{0.06477} & \multirow{4}{*}{74.36\%} & \multirow{4}{*}{7.5} & 73.18          & $\downarrow$2.60          \\
Slimming                                    &                        &                          &                          &                         & & 91.09          & $\downarrow$1.89          &                        &                          &                          &                         & & 73.19          & $\downarrow$2.59          \\
CPSE                                              &                        &                          &                          &                          & &91.28          & $\downarrow$1.70          &                        &                          &                          &                          && 73.22          & $\downarrow$2.56          \\
\textbf{CPSCA}                                          &                        &                          &                          &                          && \textbf{91.31} & \textbf{$\downarrow$1.67} &                        &                          &                          &                         & & \textbf{73.34} & \textbf{$\downarrow$2.44} \\ \hline
$\ell_{1}$-norm                                               & \multirow{4}{*}{0.04M} & \multirow{4}{*}{95.35\%} & \multirow{4}{*}{0.01351} & \multirow{4}{*}{94.65\%} & \multirow{4}{*}{1.5} & 89.32          & $\downarrow$3.66          & \multirow{4}{*}{0.05M} & \multirow{4}{*}{94.19\%} & \multirow{4}{*}{0.01352} & \multirow{4}{*}{94.65\%} & \multirow{4}{*}{1.6} & 71.23          & $\downarrow$4.55          \\
Slimming                                    &                        &                          &                          &                         & & 89.17          & $\downarrow$3.81          &                        &                          &                          &                        &  & 71.57          & $\downarrow$4.21          \\
CPSE                                              &                        &                          &                          &                         & & 88.96          & $\downarrow$4.02          &                        &                          &                          &                        &  & 71.86          & $\downarrow$3.92          \\
\textbf{CPSCA}                                          &                        &                          &                          &                          && \textbf{90.18} & \textbf{$\downarrow$2.80} &                        &                          &                          &                         & & \textbf{72.61} & \textbf{$\downarrow$3.17} \\ \hline
\end{tabular}
\end{table*}

Based on the results, we can make several observations. Firstly, the networks with SCA outperform all the counterparts in terms of inference accuracy across all four architectures and both datasets. The performance improvement verifies that SCA is an effective attention module, benefiting from the combination of both attentions as well as the adoption of new pooling strategy and grouping operations. Secondly, the results show that SCA can efficiently raise predictive accuracy with negligible extra parameters and computation overhead. Though both SCA and CBAM exploit both spatial and channel-wise attention, CBAM actually incurs noticeable overheads but can't achieve accuracy as high as that of SCA. The reason is that, the grouping operation in the spatial submodule and the replacement of the FC layers with GN operation in the channel submodule can greatly reduce both the computation overhead and the redundant parameters of SCA. Thirdly, SCA is shown to be more capable of helping the shallower models (e.g., VGG16 and ResNet56) other than the deeper ones (e.g., VGG19 and ResNet110) to boost their representation power \cite{woo2018cbam} and improve their inference performance.

\subsection{Comparisons with State-of-the-art Pruning Methods}

We have verified that our SCA achieves the best performance among all the presented attention modules. As a result, the scale values generated by SCA are more convincing, and can better describe the important levels of different channels. In this part, we compare the CPSCA pruning approach with the representative pruning schemes, including $\ell_{1}$-norm \cite{li2016pruning} and Slimming \cite{liu2017learning}. Additionally, we also apply the most classic attention module SENet for pruning \cite{song2018channel} and name the approach CPSE (Channl Pruning guided by SENet), and use the result as a comparative reference. Compared with existing pruning algorithms, our CPSCA approach needs to embed the SCA modules into the original network before pruning, and remove all of them after obtaining the attention statistics. Thus, it may be more laborious in operation. Table \ref{tb4} and Table \ref{tb5} compare the obtained accuracy of different solutions under the same pruning ratios on CIFAR-10 and CIFAR-100, respectively. The column of "Time" in the tables is the average inference time for one image on the NVIDIA Jetson Nano edge platform.

From the results in Table \ref{tb4} and Table \ref{tb5}, our CPSCA approach consistently outperforms the other state-of-the-art pruning methods, which demonstrates the effectiveness of CPSCA on the two datasets. It is also worth mentioning that when the pruning ratio is relatively small, CPSCA may have a higher accuracy than that of the original model. For example, CPSCA improves the accuracy of VGG16 by 0.41\% on CIFAR-10 dataset when the pruning ratio is 32.25\%, and improves the accuracy of ResNet56 by 0.09\% on CIFAR-100 dataset when the pruning ratio is 47.67\%. We hypothesize that the risk of model overfitting could be partially mitigated by pruning some unimportant channels. Though similar results are also reported for the other counterparts, these schemes are generally more sensitive to the increment of pruning ratio than CPSCA. As another extreme, when the pruning ratio is relatively very high (e.g., 95\%), CPSCA still maintains competitive accuracy. The accuracy degradation in SCA is up to 1.65\% for VGG16 and up to 3.17\% for ResNet56 respectively on both datasets, much less than the corresponding results of the other methods. In particular, our CPSCA pruning algorithm can significantly accelerate the inference computation on the real edge device. All these observations clearly demonstrate that it is beneficial to prune channels with the guidance of our SCA module.

\section{Conclusions and Future Work}
In this work, we have proposed a new channel pruning approach called Channel Pruning guided by Spatial and Channel Attention (CPSCA) to compress DNN models in edge computing. The core idea of CPSCA is pruning channels with least importance based on the attention statistics provided by a new attention module SCA. By combining spatial and channel attention as a whole, the SCA module can not only enhance the representation power of DNN model, but also reveal the channels' existence to inference performance. Comprehensive experiments on two benchmark datasets verify the effectiveness of the SCA module and the CPSCA approach, as compared to other state-of-the-art solutions.

For future work, we will investigate the performance of our CPSCA with other popular datasets (e.g., ImageNet) and more DNN models for edge intelligence. We also plan to combine this approach with other model compression strategies (e.g., quantization) and other edge intelligence technologies (e.g, edge-cloud collaboration) to further reduce model size and inference cost.

\section*{Acknowledgments}
This work is supported by the Beijing Municipal Natural Science Foundation under Grant L191019, the Open Project of Key Laboratory of Industrial Internet of Things \& Networked Control, Ministry of Education under Grant 2019FF03, the Open Project of Beijing Intelligent Logistics System Collaborative Innovation Center under Grant BILSCIC-2019KF-10, the CERNET Innovation Project under Grant NGII20190308, the Research Base Project of Beijing Municipal Social Science Foundation under Grant 18JDGLB026 and the Science and Technique General Program of Beijing Municipal Commission of Education under Grant KM201910037003.

\bibliographystyle{elsarticle-num-names}
\bibliography{ref}

\begin{wrapfigure}{l}{25mm}
    \includegraphics[width=1in,height=1.25in,clip,keepaspectratio]{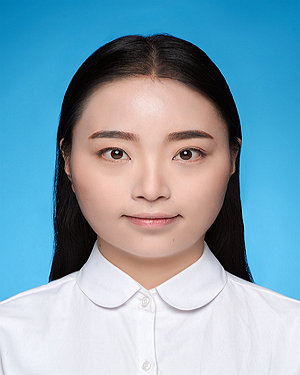}
  \end{wrapfigure}\par
  \footnotesize{
  \textbf{~\\
  Mengran Liu} received her B.S. degree from the Department of Internet of Things Engineering from TianGong University in 2018. Currently, she is a graduate student in the School of Computer and Information Technology at Beijing Jiaotong University. Her main current research interests include edge computing and Internet of Things.\par}
~ \\
~ \\
~ \\
~ \\
~ \\
~ \\
 \begin{wrapfigure}{l}{25mm}
    \includegraphics[width=1in,height=1.25in,clip,keepaspectratio]{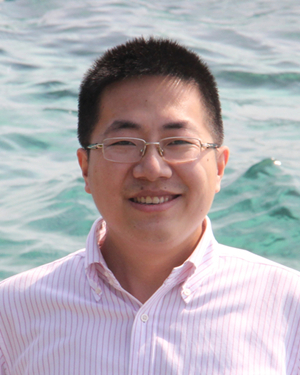}
  \end{wrapfigure}\par
  \footnotesize{
  \textbf{~\\
  Weiwei Fang} received the B.S. degree from Hefei University of Technology, Hefei, China, and the Ph.D. degree from Beihang University, Beijing, China, in 2003 and 2010, respectively. He is currently an Associate Professor with the School of Computer and Information Technology, Beijing Jiaotong University, Beijing, China. His research interests include edge computing, cloud computing, and Internet of Things. He has published over 60 papers in journals, international conferences/workshops.\par}
~ \\
~ \\
~ \\
~ \\
 \begin{wrapfigure}{l}{25mm}
    \includegraphics[width=1in,height=1.25in,clip,keepaspectratio]{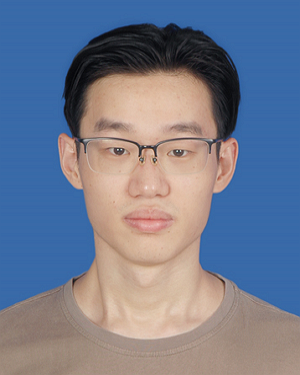}
  \end{wrapfigure}\par
  \footnotesize{
  \textbf{~\\
  Xiaodong Ma} received his B.S. degree from the School of Software Engineering from Zhengzhou University in 2020. Currently, he is a graduate student in the School of Computer and Information Technology at Beijing Jiaotong University. His current research interests include edge computing and distributed machine learning.\par}
~ \\
~ \\
~ \\
~ \\
~ \\
~ \\
~ \\
 \begin{wrapfigure}{l}{25mm}
    \includegraphics[width=1in,height=1.25in,clip,keepaspectratio]{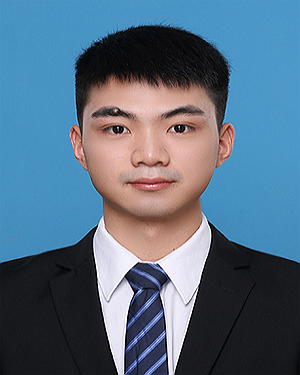}
  \end{wrapfigure}\par
  \footnotesize{
  \textbf{~\\
  Wenyuan Xu} received his B.S. degree from the School of Software Engineering from Zhengzhou University in 2019. Currently, he is a graduate student in the School of Computer and Information Technology at Beijing Jiaotong University. His current research interests include edge computing and DNN model compression.\par}
~ \\
~ \\
~ \\
~ \\
~ \\
~ \\
~ \\
 \begin{wrapfigure}{l}{25mm}
    \includegraphics[width=1in,height=1.25in,clip,keepaspectratio]{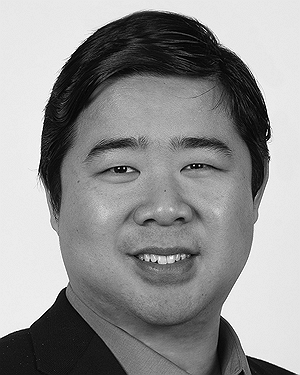}
  \end{wrapfigure}\par
  \footnotesize{
  \textbf{~\\
  Naixue Xiong} received the Ph.D. degree in sensor system engineering from Wuhan University and in dependable sensor networks from the Japan Advanced Institute of Science and Technology. He was with Georgia State University, Wentworth Technology Institution, and Colorado Technical University (as a Full Professor for five years) about ten years. He is currently a Professor with the College of Intelligence and Computing, Tianjin University, China. He has published over 300 international journal articles and over 100 international conference papers. Some of his works were published in the IEEE JOURNAL ON SELECTED AREAS IN COMMUNICATIONS, IEEE or ACM transactions, ACM Sigcomm workshop, IEEE INFOCOM, ICDCS, and IPDPS. His research interests include cloud computing, security and dependability, parallel and distributed computing, networks, and optimization theory. Dr. Xiong has been a Senior member of the IEEE Computer Society, since 2012. He has received the Best Paper Award in the 10th IEEE International Conference on High Performance Computing and Communications (HPCC-08) and the Best student Paper Award in the 28th North American Fuzzy Information Processing Society Annual Conference (NAFIPS2009).\par}
~ \\
\begin{wrapfigure}{l}{25mm}
    \includegraphics[width=1in,height=1.25in,clip,keepaspectratio]{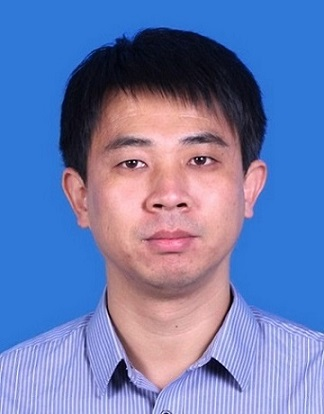}
  \end{wrapfigure}\par
  \footnotesize{
  \textbf{~\\
  Yi Ding} received the Ph.D. degree from Beihang University, Beijing, China, in 2014. Currently, he is an Assistant Professor in the School of Information at Beijing Wuzi University. His current research interests include edge computing and blockchain systems.\par}

\end{document}